\documentclass[10pt,twocolumn,letterpaper]{article}

\usepackage{iccv}
\usepackage{times}
\usepackage{epsfig}
\usepackage{graphicx}
\usepackage{amsmath}
\usepackage{amssymb}
\usepackage{multirow}
\usepackage{float}

\usepackage[pagebackref=true,breaklinks=true,letterpaper=true,colorlinks,bookmarks=false]{hyperref}

\iccvfinalcopy 

\def\iccvPaperID{1286} 

\DeclareMathOperator*{\argmin}{argmin}

\ificcvfinal\fi
\begin{document}

\title{A Two-Streamed Network for Estimating Fine-Scaled \\
Depth Maps from Single RGB Images}

\author{Jun Li\textsuperscript{1,2}, Reinhard Klein\textsuperscript{1} and Angela Yao\textsuperscript{1}\\
\textsuperscript{1}University of Bonn, 
\textsuperscript{2}National University of Defense Technology\\
{\tt\small {lij, rk, yao}@cs.uni-bonn.de}
}

\maketitle

\begin{abstract}
Estimating depth from a single RGB image is an ill-posed and inherently ambiguous problem. State-of-the-art deep learning methods can now estimate accurate 2D depth maps, but when the maps are projected into 3D, they lack local detail and are often highly distorted. We propose a fast-to-train two-streamed CNN that predicts depth and depth gradients, which are then fused together into an accurate and detailed depth map. We also define a novel set loss over multiple images; by regularizing the estimation between a common set of images, the network is less prone to over-fitting and achieves better accuracy than competing methods.  Experiments on the NYU Depth v2 dataset shows that our depth predictions are competitive with state-of-the-art and lead to faithful 3D projections.
\end{abstract}












\section{Introduction}
Estimating depth for common indoor scenes 
from monocular RGB images 
has widespread applications in scene understanding, depth-aware image editing or re-rendering, 3D modelling, robotics, \etc.  Given a single RGB image as input, the goal is to predict a dense depth map for each pixel.  Inferring the underlying depth is an ill-posed and inherently ambiguous problem. In particular, indoor scenes have large texture and structural variations, heavy object occlusions and rich geometric detailing, all of which contributes to the difficulty of accurate depth estimation.

\begin{figure*}[t!]
    \centering
    \includegraphics[width=1.0\textwidth]{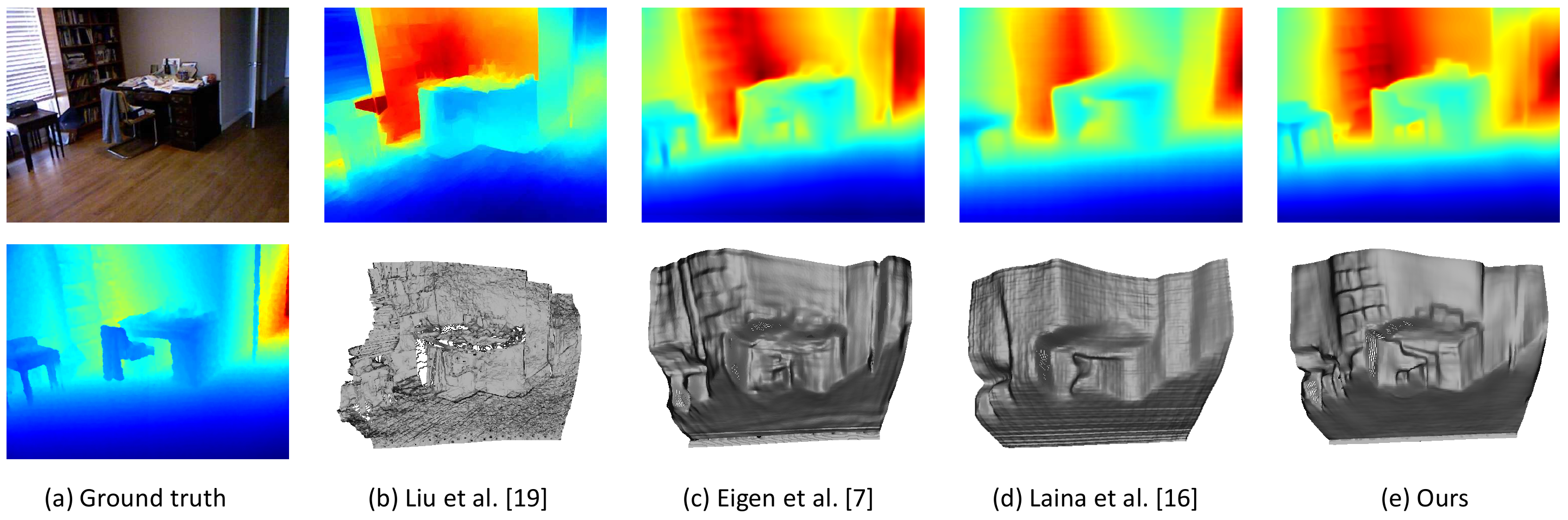}
    \label{fig:example_1}
    \vspace{-0.4cm}
    \caption{3D projects from estimated depth maps of state-of-the-art methods (b-d), along with ground truth (a). Our estimated depth maps(e) are more accurate than state-of-the-art methods with fine-scaled details. Note that colours values of each depth map are individually scaled.}
    \vspace{-0.4cm}
\end{figure*}

The use of convolutional neural networks (CNNs) has greatly improved the accuracy of depth estimation techniques~\cite{David15, Garg16, Kim16, laina2016deeper, Li15,Liu15, Roy15,Wang15}. Rather than coarsely approximating the depth of large structures such as walls and ceilings, state-of-the-art networks~\cite{David15,laina2016deeper} benefit from using pre-trained CNNs and can capture fine-scaled items such as furniture and home accessories. The pinnacle of success for depth estimation is the ability to generate realistic and accurate 3D scene reconstructions from the estimated depths.  Faithful reconstructions should be rich with local structure; detailing becomes especially important in applications derived from the reconstructions such as object recognition and depth-aware image re-rendering and or editing.  Despite the impressive evaluation scores of recent works~\cite{David15,laina2016deeper} however, the estimated depth maps still suffer from artifacts at finer scales and have unsatisfactory alignments between surfaces. These distortions are especially prominent when projected into 3D (see Figure~\ref{fig:example_1}).

Other CNN-based end-to-end applications 
such as semantic segmentation~\cite{chen2016deeplab,Long15} and normal estimation~\cite{bansal2016marr,David15} face similar challenges of preserving local details.  The repeated convolution and pooling operations are critical for capturing the entire image extent, but simultaneously shrink resolution and degrade the detailing.  While up-convolution and feature map concatenation strategies~\cite{Fischer15,laina2016deeper,Long15,ronneberger2015u} have been proposed to improve resolution, output map boundaries often still fail to align with image boundaries.  As such, optimization measures like bilateral filtering~\cite{BarronPoole2016} or  CRFs~\cite{chen2016deeplab} yield further improvements.

It is with the goal of preserving detailing that we motivate our work on depth estimation.  We want to benefit from the accuracy of CNNs, but avoid the degradation of resolution and detailing.  First, we ensure network accuracy and generalization capability by introducing a novel set image loss. This loss is defined jointly over multiple images, where each image is a transformed version of an original image via standard data augmentation techniques. The set loss considers not only the accuracy of each transformed image's output depth, but also has a regularization term to minimize prediction differences within the set.  Adding this regularizer greatly improves the depth accuracy and reduces RMS error by approximately $5\%$.  As similar data augmentation approaches are also used in other end-to-end frameworks, \eg for semantic segmentation and normal estimation, we believe that the benefits of the set loss will carry over to these applications as well.

\begin{figure*}[!h]
    \centering
    \includegraphics[width=17cm]{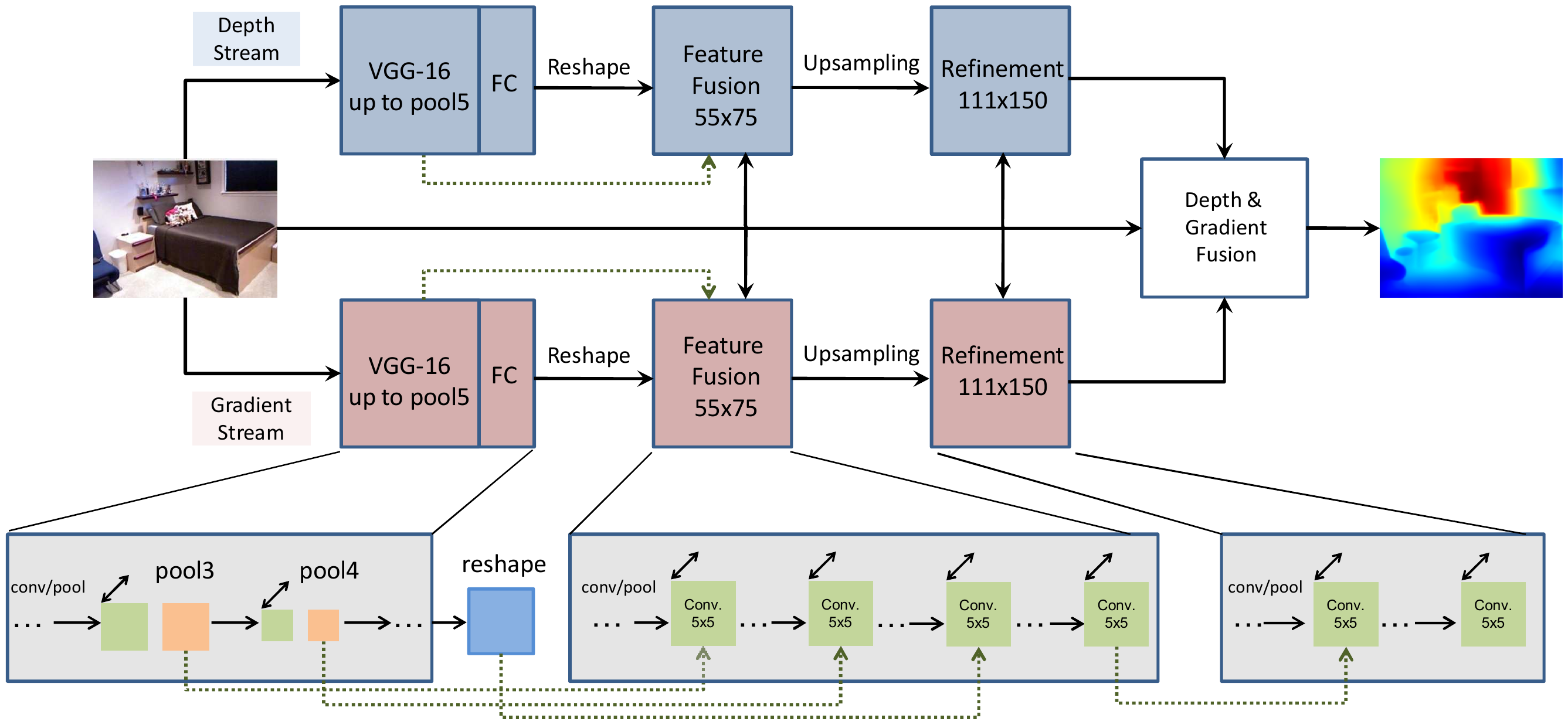}

 \vspace{0.2cm}
    \caption{
    Our two-streamed depth estimation network architecture; the top stream (blue) estimates depth while the bottom (pink) estimates depth gradients. The dotted lines represent features fused from the VGG convolutional layers (see Section~\ref{sec:NetArch}). The depth and depth gradients are then combined either via further convolution layers or directly with an optimization enforcing consistency between the depth and depth gradients.
    Figure is best viewed in colour.
     }\label{fig:arch}    
\end{figure*}

We capture scene detailing by considering information contained in depth gradients. We postulate that local structure can be better encoded with first-order derivative terms than the absolute depth values.  Perceptually, it is sharp edges and corners which define an object and make it recognizable, rather than (correct) depth values (compare in Figure~\ref{fig:results}). 
As such, we think it's better to represent a scene with both depth and depth gradients, and propose a fast-to-train two-streamed CNN to regress the depth and depth gradients (see Figure~\ref{fig:arch}).  In addition, we propose two possibilities for fusing the depth and depth gradients, one via a CNN, to allow for end-to-end training, and one via direct optimization. We summarize our contributions as follows:

\begin{itemize}
\item A novel set image loss with a regularizer that minimizes differences in estimated depth of related images; this loss makes better use of augmented data and promotes stronger network generalization, resulting in higher estimation accuracy.

\item A joint representation of the 2.5D scene with depth and depth gradients; this representation captures local structures and fine detailing and is learned with a two-streamed network.

\item Two methods for fusing depth and depth gradients into a final depth output, one via CNNs for end-to-end training and one via direct optimization; both methods yield depth maps which, when projected into 3D, have less distortion and are richer with structure and object detailing than competing state-of-the-art.
\end{itemize}


Representing the scene with with both depth and depth gradients is redundant, as one can be derived from the other. We show, however, that this redundancy offers explicit consideration for local detailing that is otherwise lost in the standard Euclidean loss on depth alone and/or with a simple consistency constraint in the loss. Our final depth output is accurate and clean with local detailing, with fewer artifacts than competing methods when projected into 3D.

\section{Related Work}
Depth estimation 
is a rich field of study and we discuss only the monocular methods.  
A key strategy in early works for handling depth ambiguity was to use strong assumptions and prior knowledge. For example, Saxena \etal~\cite{Saxena05,Saxena08} devised a multi-scale MRF, but assumed that all scenes were horizontally aligned with the ground plane. 
Hoiem \etal~\cite{Hoiem05}, instead of predicting depth explicitly, estimated geometric structures for major image regions and composed simple 3D models to represent the scene. 

Once RGB-D data could be collected from laser or depth cameras on a large scale, it became feasible to apply data-driven learning-based approaches
~\cite{Karsch14, Liu14, Saxena05,Saxena08,Zhuo15}.  
Karsch \etal~\cite{Karsch14} proposed a non-parametric method to transfer depth from aligned exemplars and formulated depth estimation as an optimization problem with smoothness constraints. Liu \etal~\cite{Liu14} modelled image regions as super-pixels and used discrete-continuous optimization for depth estimation and later integrated  mid-level region features and global scene layout~\cite{Zhuo15}. Others tried to improve depth estimations by exploiting semantic labels~\cite{Hane15,Ladick14,Liu10}.
With hand-crafted features, however, the inferred depth maps are coarse and only approximate the global layout of a scene.  Furthermore, they lack the finer details necessary for many applications in computer vision and graphics. 


Deep learning has proven to be highly effective for depth estimation~\cite{David15,Garg16,Kim16,Li15,Liu15,Wang15,Roy15}. Liu \etal~\cite{Liu15} combined CNNs and CRFs in a unified framework to learn unary and pairwise potentials with CNNs.  They predicted depth at a superpixel level which works well for preserving edges, but when projected to 3D, suffers from distortions and artifacts, as each superpixel region retains the same or very similar depth after an in-painting post-processing.  

More recent methods~\cite{Chakrabarti16,David15,laina2016deeper}
have the harnessed the power of pre-trained CNNs in the form of fully convolutional networks~\cite{Long15}. 
The convolutional layers from networks such as VGG~\cite{Simonyan15} and ResNet~\cite{he2016deep} are fine-tuned, while the fully connected layers are re-learned from scratch to encode a spatial feature mapping of the scene. The learned map, however, is at a much lower resolution than the original input. To recover a high-resolution depth image, the feature mapping is then up-sampled~\cite{Chakrabarti16,David15} or passed through up-convolution blocks~\cite{laina2016deeper}.  Our network architecture follows a similar fully convolutional approach, and increases resolution via up-sampling.  In addition, we add skip connections between the up-sampling blocks to better leverage intermediate outputs.

\section{Learning}\label{sec:Loss}
\subsection{Network Architecture}\label{sec:NetArch}
Our network architecture, shown in Figure~\ref{fig:arch}, follows a two-stream model; one stream regresses depth and the other depth gradients, both from an RGB input image. 
The two streams follow the same format: an image parsing block, flowed by a feature fusion block and finally a refinement block. The image parsing block consists of the convolutional layers of VGG-16 (up to \emph{pool5}) and two fully connected layers. 
The output from the second fully connected layer is then reshaped into a $55\!\times\!75\!\times\!D$ feature map to be passed onto the feature fusion block, where $D\!=\!1$ for the depth stream and $D\!=\!2$ for the gradient stream.  In place of VGG-16, other pre-trained networks can be used as well for the image parsing block, \eg VGG-19 or ResNet.

The feature fusion block consists of one $9\!\times\!9$ convolution and pooling, followed by eight successive  $5\!\times\!5$ convolutions without pooling. It takes as input a down-sampled RGB image and then fuses together features from the VGG convolutional layers and the image parsing block output.  Specifically, the features maps from VGG \emph{pool3} and \emph{pool4}
are fused at the input to the second and fourth convolutional layers respectively, while the output of the image parsing block is fused at the input to the sixth convolutional layer, all with skip layer connections. The skip connections for the VGG features have a $5\!\times\!5$ convolution and a 2x or 4x up-sampling to match the working $55\!\times\!75$ feature map size; the skip connection from the image parsing block is a simple concatenation. As noted by other image-to-image mapping works~\cite{pix2pix2016,Long15,ronneberger2015u}, the skip connections provide a convenient way to share hierarchical information and we find that this also results in much faster network training convergence. The output of the feature fusion block is a coarse $55\!\times\!75\!\times\!D$ depth or depth gradient map.

The refinement block, similar to the feature fusion block, consists of one $9\!\times\!9$ convolution and pooling and five $5\!\times\!5$ convolutions without pooling. It takes as input a down-sampled RGB image and then fuses together a bilinearly up-sampled output of the feature fusion block 
via a skip connection (concatenation) to the third convolutional layer.  The working map size in this block is $111\!\times\!150$, with the output being depth or gradient maps at this higher resolution. 

The depth and gradient fusion block brings together the depth and depth gradient estimates from the two separate streams into a single coherent depth estimate.  We propose two possibilities, one with convolutional processing in an end-to-end network, and one via a numerical optimization.  The two methods are explained in detail in Section~\ref{sec:fusion}. We refer the reader to the supplementary material for specifics on the layers, filter sizes, and learning rates.  

\subsection{Set Image Loss}
For many machine learning problems, it has become standard practice to augment the training set with transformed versions of the original training samples.  By learning with the augmented set, the resulting classifier or regressor should be more robust to these variations.  The loss is applied over some batch of the augmented set, where transformed samples are treated as standard training samples. However, there are strong relations between original and transformed samples that can be further leveraged during training. For example, a sample image that is re-coloured to approximate different lighting conditions should have exactly the same depth estimate as the original. A flipped sample will also have the same depth estimate as the original after un-flipping the output and so on.

Based on this observation, we formulate the set loss as follows. We start by  defining the pixel-wise $l_2$ difference between two depth maps $D_1$ and $D_2$:

\begin{equation}
l_2(D_1,D_2) = \frac{1}{n}\sum_{p}(D_1^p-D_2^p)^{2},
\label{eq:commonLoss}
\end{equation}
\noindent where $p$ is a pixel index, to be summed over $n$ valid depth pixels\footnote{Invalid pixels are parts of the image with missing ground truth depth; such holes are typical of Kinect images and not considered in the loss.}.  A simple error measure comparing an estimated depth map $D_i$ and its corresponding ground truth $D_{\text{gt}_i}$ can then be defined as $l_2(D_i, D_{\text{gt}_i})$.

Now consider an image set  $\{I,f_{1}(I),f_{2}(I)...f_{N-1}(I)\}$, of size $N$, where $I$ is the original image and $f$ are the data augmentation transformations such as colour adjustment, flipping, rotation, skew, \etc. For a set of images, the set loss $L_{\text{set}}$ is given by
\vspace{-0.1cm}
\begin{equation}\label{eq:jointloss}
    L_{\text{set}} = L_{\text{single}} + \lambda\cdot{\Omega_{\text{set}}},
\end{equation}

\noindent which considers the estimation error of each image in the set as independent samples in $L_{\text{single}}$, along with a regularization term $\Omega_{\text{set}}$ based on the entire set of transformed images, with $\lambda$ as a weighting parameter between the two. 

More specifically, $L_{\text{single}}$ is simply the the estimation error of each (augmented) image considered individually, \ie
\begin{equation}\label{eq:gtLoss}
L_{\text{single}} = \frac{1}{N}\sum_{i=1}^{N}l_2(D_{i}, D_{\text{gt}_{i}}).
\end{equation}

\noindent The regularization term $\Omega_{\text{set}}$ is defined as the $l_2$ difference between estimates within an image set:

\begin{equation}\label{eq:setLoss}
\Omega_{\text{set}} = \frac{2}{N (N-1)}\sum_{i=1}^N\sum_{j>i}^{N}l_2(D_{i}, g_{ij}(D_{j})).
\end{equation}


\noindent Here, $D_{i}$ and $D_{j}$ are the depth estimates of any two images $I_{i}$ and $I_{j}$ in the image set.  $g$ is a mapping between their transformations, \ie $g_{ij}(\cdot)  = f_i(f^{-1}_{j}(\cdot))$.  For all $f$ which are spatial transformations, $g$ should realign the two estimated depth maps into a common reference frame for comparison. For colour transformations, $f$ and $f^{-1}$ are simply identity functions. $\Omega_{\text{set}}$ acts as a consistency measure which encourages all the depth maps of an image set to be the same after accounting for the transformations.  This regularization term has loose connections to tangent propagation~\cite{Simard92}, which also has the aim of encouraging invariance to input transformations. Instead of explicit computing the tangent vectors, we transform augmented samples back into a common reference before doing standard back-propagation. 

\subsection{Depth and Depth Gradient Estimation}\label{sec:fusion}
To learn the network in the depth stream, we use our proposed set loss as defined in Equations~\ref{eq:jointloss} to~\ref{eq:setLoss}.  For learning the network in the depth gradient stream, we use the same formulation, but modify the pixel-wise difference for two gradient maps $G_1$ and $G_2$, substituting $l_{2g}$ for $l_2$:

\vspace{-0.1cm}
\begin{equation}
l_{2g}(G_1,G_2) = \frac{1}{n}\sum_{p}(G_{1x}^p-G_{2x}^p)^{2}+(G_{1y}^p-G_{2y}^p)^{2},
\label{eq:commonGradLoss}
\end{equation}
where $n$ is the number of valid depth pixels and $G_{x}^p$ and $G_{y}^p$ the $X$ and $Y$ gradients at pixel $p$ respectively.  

\paragraph{Fusion in an End-to-end Network}
We propose two possibilities for fusing the outputs of the depth and gradient streams into a final depth output. The first is via a combination block, with the same architecture as the refinement block. It takes as input the RGB image and fuses together the depth estimates and gradient estimates via skip connections (concatenations) as inputs to the third convolutional layer. We use the following combined loss $L_{\text{comb}}$ that maintains depth accuracy and gradient consistency:

\begin{equation}
\begin{aligned}
L_{\text{comb}} = L_{\text{set}}+\frac{1}{N}\sum_{i=1}l_{2g}(\nabla{D_i}, G_{\text{est}_{i}}),
\label{eq:combinationLoss}
\end{aligned}
\end{equation}
where $\nabla{D_i}$ indicates application of the gradient operator on depth map $D_i$. In this combined loss, the first term $L_{\text{set}}$ is based only on depth, while the second term enforces consistency between the gradient of the final depth and estimated gradients with the same $l_{2g}$ pixel-wise difference from Equation~\ref{eq:commonGradLoss}.



\paragraph{Fusion via Optimization}
Alternatively, as optimization measures have also shown to be highly effective in improving output map detailing~\cite{BarronPoole2016,chen2016deeplab}, we directly estimate an optimal depth $D^*$ based on the following minimization:

\begin{equation}
\begin{aligned}
\vspace{-0.3cm}
D^* = & \argmin_{D} \sum_{p = 1}^n \phi(D^p-D^p_{\text{est}})+\\
&\omega\sum_{p}^n \big[\phi(\nabla_{x}D^p-G^p_x)+\phi(\nabla_{y}D^p-G^p_y)\big],\\
\label{eq:optLoss}
\end{aligned}
\end{equation}

\noindent where $D_{est}$ is the estimated depth and $G_{x}$ and $G_{y}$ are the estimated gradients in $x$ and $y$. $\phi(x)$ acts as a robust $L_1$ measure, \ie $\phi(x)\!=\!\sqrt{x^2+\epsilon}, \,\, \epsilon\!\!=\!\!10^{-4}$; $\nabla_{x}$, $\nabla_{y}$ are $x$, $y$ gradient operators on depth $D$ (we use filters $[-1,0,1]$, $[-1,0, 1]^\intercal$) and $p$ is the pixel index summed over the $n$ valid pixels.  We solve for $D$ with iteratively re-weighted least squares using the implementation provided by Karsch~\cite{Karsch14}.

\subsection{Training Strategy}
We apply the same implementation for the networks in both the depth and the gradient streams.  With the exception of the VGG convolutional layers, the fully connected layers and all layers in the feature fusion, refinement and combination block are initialized randomly.

The two streams are initially trained individually, each with a two-step procedure.  First, the image parsing and feature fusion blocks are trained with a loss on the depth and depth gradients.  These blocks are then fixed, while in the refinement blocks are trained in a second step.  For each step, the same set loss with the appropriate pixel differences (see Equations~\ref{eq:jointloss},~\ref{eq:commonLoss},~\ref{eq:commonGradLoss}) is used, albeit with different map resolutions ($55\!\times\!75$ after feature fusion, $111\!\times\!75$ after refinement). Training ends here if the optimization-based fusion is applied.  Otherwise, for the end-to-end network fusion, the image parsing, feature fusion and refinement blocks are fixed while the fusion block is trained using the combined loss (Equation~\ref{eq:combinationLoss}).  A final fine-tuning is applied to all the blocks of both streams jointly based on this combined loss.  

The network is fast to train; only 2 to 3 epochs are required for convergence (see Figure~\ref{fig:convergence}). During the first 20K iterations, we stabilize training with gradient clipping.  For fast convergence, we use a batch size of $1$; note that because our loss is defined over an image set, a batch size of $1$ is effectively a mini-batch of size $N$, depending on the number of image transformations used.   

The regularization constants $\lambda$ and $\omega$ are set to 1 and 10 respectively. Preliminary experiments showed that different values of $\lambda$ which  controls the extent of the set image regularizer does not affect the resulting accuracy, although a larger $\lambda$ does slow down network convergence. $\omega$, controlling the extent of the gradients in the gradient optimization, was set by a validation set; a larger $\omega$ over-emphasizes artifacts in the gradient estimates and leads to less accurate depth maps.  

\section{Experimentation}

\subsection{Dataset \& Evaluation}
We use the NYU Depth v2 dataset~\cite{Silberman12} with the standard scene splits; from the 249 training scenes, we extract $\sim$220k training images. RGB images are down-sampled by half and then cropped to $232\!\times\!310$ to remove  blank boundaries after alignment with depth images. Depths are converted to log-scale while gradients are kept in a linear scale.

\begin{table*}[!t]
\centering 
\begin{tabular}{| l | c | c c c | c c c|}
\hline
\multicolumn{1}{|c|}{
Method} & Base Network & rel & log10 & rms & $\delta<1.25$ & $\delta<1.25^2$ & $\delta<1.25^3$\\
\hline
Karsch \etal~\cite{Karsch14} & - & 0.35 & 0.131 & 1.2 & - & - & -\\
Liu \etal~\cite{Liu14} & - & 0.335 & 0.127 & 1.06 & - & - & -\\
Ladicky \etal~\cite{Ladick14} & - & - & - & - & 0.542 & 0.829 & 0.941\\
Li \etal~\cite{Li15} & - & 0.232 & 0.094 & 0.821 & 0.621 & 0.886 & 0.968\\
Liu \etal~\cite{Liu15} & - & 0.213 & 0.087 & 0.759 & 0.650 & 0.906 & 0.976\\
Eigen \etal~\cite{David15} & VGG-16 & 0.158 & - & 0.641 & 0.769 & 0.950 & 0.988\\
Laina \etal~\cite{laina2016deeper} & VGG-16 & 0.194 & 0.083 & 0.790 & 0.629 & 0.889 & 0.971\\
Chakrabarti \etal~\cite{Chakrabarti16} & VGG-19 & 0.149 & - & 0.620 & 0.806 & 0.958 & 0.987\\
Laina \etal~\cite{laina2016deeper}   & ResNet-50 & 0.127 & 0.055 & 0.573 & 0.811 &  0.953 & 0.988 \\
\hline
 $L_{\text{single}}$, depth only & VGG-16 & 0.161 & 0.068 & 0.640 & 0.765 & 0.950 & 0.988\\
$L_{\text{set}}$, depth only & VGG-16 & 0.153 & 0.065 & 0.617 & 0.786 & 0.954 & 0.988\\
$L_{\text{set}}$, depth + bilateral filtering & VGG-16 & 0.152 & 0.065 & 0.621 & 0.785 & 0.954 & 0.988\\
$L_{\text{set}}$, depth + gradients, end-to-end & VGG-16 & 0.153 & 0.064 & 0.615 & 0.788 & 0.954 & 0.988\\
$L_{\text{set}}$, depth + gradients, optimization & VGG-16 & 0.152 & 0.064 & 0.611 & 0.789 & 0.955 & 0.988\\
$L_{\text{set}}$, depth + gradients, optimization & VGG-19 & 0.146 & 0.063 & 0.617 & 0.795 & 0.958 & 0.991\\
$L_{\text{set}}$, depth + gradients, optimization & ResNet-50 & 0.143 & 0.063 & 0.635 & 0.788 &0.958 & 0.991\\
\hline
\end{tabular}
\vspace{0.2cm}
\caption{Accuracy of depth estimates on the NYU Depth v2 dataset, as compared with state of the art. Smaller values on $\text{rel}$, $\log_{10}$ and $\text{rms}$ error are better; higher values on $\delta < \text{threshold}$ are better.  
}
\label{table:accuracy}
\vspace{-0.2cm}
\end{table*}

We evaluate our proposed network and the two fusion methods on the 654 NYU Depth v2~\cite{Silberman12} test images. Since our depth output is $111\times150$ and a lower resolution than the original NYUDepth images, we bilinearly up-sample our depth map (4x) and fill in missing borders with a cross-bilateral filter, similar to previous methods~\cite{David15,Li15,Liu15,Liu14}.
We evaluate our predictions in the valid Kinect depth projection area, using the same measures as previous work: 
\begin{itemize}
\item mean relative error (rel): $\frac{1}{T}\sum_{i}^{T}\frac{|d^{gt}_i - d_i|}{d^{gt}_i}$,
\item mean $\log_{10}$ error ($\log_{10}$): $\!\frac{1}{T}\sum_{i}^{T}|\log_{10}d^{gt}_i$ $-\log_{10}d_i|$,
\item root mean squared error (rms): $\sqrt{\frac{1}{T}\sum_{i}^{T}\left (d^{gt}_i - d_i\right )^2}$,
\item thresholded accuracy: percentage of $d_{i}$ such that $\max \left ({d^{gt}_i}/{d_i}, {d_i}/{d^{gt}_i} \right )  = \delta < \text{threshold}$.  
\end{itemize}
In each measure, $d^{gt}_i$ is the ground-truth depth, $d_i$ the estimated depth, and $T$ the total pixels in all evaluated images. Smaller values on $\text{rel}$, $\log_{10}$ and $\text{rms}$ error are better 
and higher values on percentage(\%) $\delta < \text{threshold}$ are better.

We make a qualitative comparison by projecting the estimated 2D depth maps into a 3D point cloud.  The 3D projections are computed using the Kinect camera projection matrix and the resulting point cloud is rendered with lighting.  Representative samples are shown in Figure~\ref{fig:results}; the reader is referred to the Supplementary Material for more results over the test set.

\subsection{Depth Estimation Baselines}
The accuracy of our depth estimates is compared with other methods in Table~\ref{table:accuracy}. 
We consider as our baseline the accuracy of only the depth stream with a VGG-16 base network, without adding gradients ($L_{\text{single}}$, depth only). This baseline already outperforms~\cite{laina2016deeper} (VGG-16) and is comparable to~\cite{David15} (VGG-16). With the set loss, however ($L_{\text{set}}$, depth only), we surpass~\cite{David15} with more accurate depth estimates, especially in terms of rms-error and thresholded accuracy with $\delta <1.25$.

Current state-of-the-art results are achieved with fully convolutional approaches~\cite{Chakrabarti16,David15,laina2016deeper}.  The general trend is that deeper base networks (VGG-19, ResNet-50 vs. VGG-16) leads to higher depth accuracy. We observe a similar trend in our results, though improvements are not always consistent.  We achieve some gains with VGG-19 over VGG-16. However, unlike~\cite{laina2016deeper}, we found little gains with ResNet-50.


\subsection{Fused Depth and Depth Gradients}
\begin{figure*}
    \centering
    \includegraphics[width=1.0\textwidth]{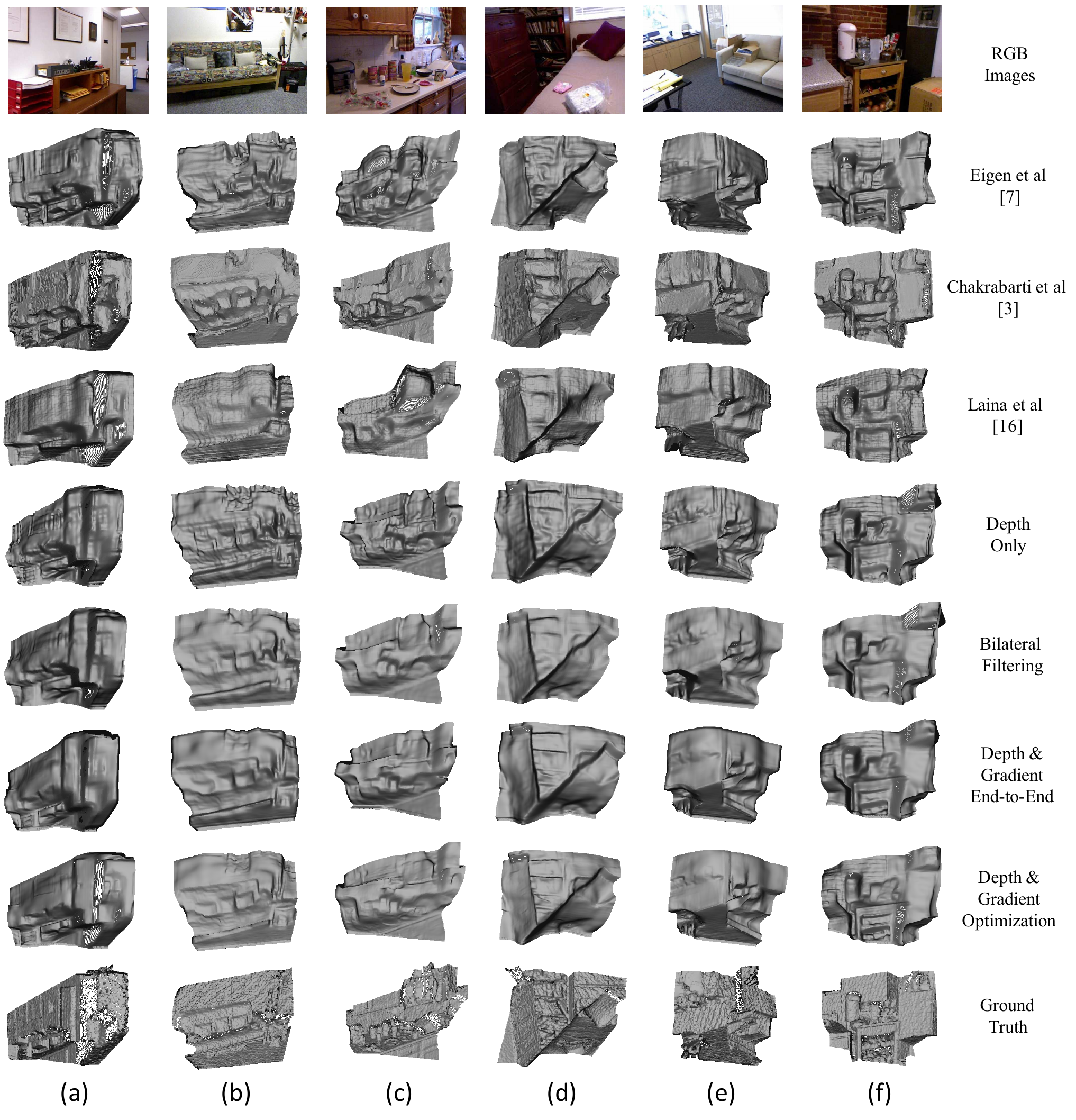}
    \vskip 0.2cm
    \begin{tabular}{| c | c c c c c c|}
    \hline
    RMS error & a & b & c & d & e & f\\
    \hline
    Eigen \etal~\cite{David15} & 0.795 & 0.154 & 0.377 & 0.204 & 0.452 & 0.253\\
    Charkrabarti \etal~\cite{Chakrabarti16} & 0.558 & 0.236 & 0.223 & 0.208 & 0.400 & 0.304\\
    Laina \etal~\cite{laina2016deeper} & 0.824 & 0.237 & 0.306 & 0.132 & 0.344 & 0.159\\
    Depth Only & 0.715 & 0.259 & 0.246 & 0.127 & 0.335 & 0.202\\
    Bilateral Filtering & 0.715 & 0.259 & 0.245 & 0.126 & 0.334 & 0.201\\
    Depth \& Gradient (End-to-End) & 0.719 & 0.261 & 0.249 & 0.130 & 0.336 & 0.192\\
    Depth \& Gradient (Optimization) & 0.715 & 0.265 & 0.248 & 0.124 & 0.338 & 0.185\\
    \hline
    \end{tabular}
    \vspace{0.3cm}
\caption{Comparison of example 3D projections with corresponding RMS errors using VGG-16 as the base network.  Please see supplementary materials for more results. The four variations of our proposed method all report similar RMS values, though differences are more noticeable in the 3D projection. From examples (b) and (f), where ~\cite{laina2016deeper} reports the lower RMS, one can see that a numerical measure such as RMS is not always indicative of estimated depth quality.
\label{fig:results}
}  

\end{figure*}

Fusing depth estimates together with depth gradients achieves similar results as the optimization, both quantitatively and qualitatively.  
When the depth maps are projected into 3D (see Figure~\ref{fig:results}), there is little difference between the two fusion methods.  When comparing with~\cite{laina2016deeper}'s ResNet-50 results, which are significantly more accurate, one sees that~\cite{laina2016deeper}'s 3D projections are more distorted. In fact, many structures, such as the shelves in Figure~\ref{fig:results}(a), the sofa in (b) or the pillows in (b,d) are unidentifiable. Furthermore, the entire projected 3D surface seems to suffer from grid-like artifacts, possibly due to their up-projection methodology. On the other hand, the projections of~\cite{Chakrabarti16} are much cleaner and detailed, even though their method also reports lower accuracy than~\cite{laina2016deeper}. As such, it is our conclusion that the current numerical evaluation measures are poor indicators of detail preservation. This is expected, since the gains from detailing have little impact on the numerical accuracy measures. Instead, differences are more salient qualitatively, especially in the 3D projection.

Compared to~\cite{David15}, our 3D projections are cleaner and smoother in the appropriate regions,~\ie walls and flat surfaces. Corners and edges are better preserved and the resulting scene is richer in detailing with finer local structures. For example, in the cluttered scenes of Figure~\ref{fig:results}(b,c),~\cite{David15} has heavy artifacts in highly textured areas,~\eg the picture on the wall of (b), the windows in (c) and in regions with strong reflections such as the cabinet in (c).  Our results are robust to these difficulties and give a more faithful 3D projections of the underlying objects in the scene.


At a first glance, one may think that jointly representing the scene with both depth and depth gradients simply has a smoothing effect.  While the fused results are definitely smoother, no smoothing operations, 2D or 3D, can recover non-existent detail.  When applying a 2D bilateral filter with 0.1m range and $10\!\times\!10$ spatial Gaussian kernel ($L_{\text{set}}$ depth + bilateral filtering), we find little difference in the numerical measures and still some losses in detailing (see  Fig.\ref{fig:results})

\subsection{Set Loss \& Data Augmentation}\label{sec:setloss}
Our proposed set image loss has a large impact in improving the estimated depth accuracy.  For the reported results in Table~\ref{table:accuracy}, we used three images in the image set: $I$, flip($I$) and colour($I$).  The flip is around the vertical axis, while the colour operation includes randomly increasing or decreasing brightness, contrast and multiplication with a random RGB value $\gamma$ $\in$ ${[0.8, 1.2]}^{3}$.  Preliminary experiments showed that adding more operations like rotation and translation did not further improve the performance so we omit them from our training. We speculate that the flip and colouring operations bring the most global variations but leave pinpointing the exact cause for future work.


\subsection{Training Convergence and Timing}
\begin{figure}[t!]
    \centering
    \includegraphics[width=0.5\textwidth]{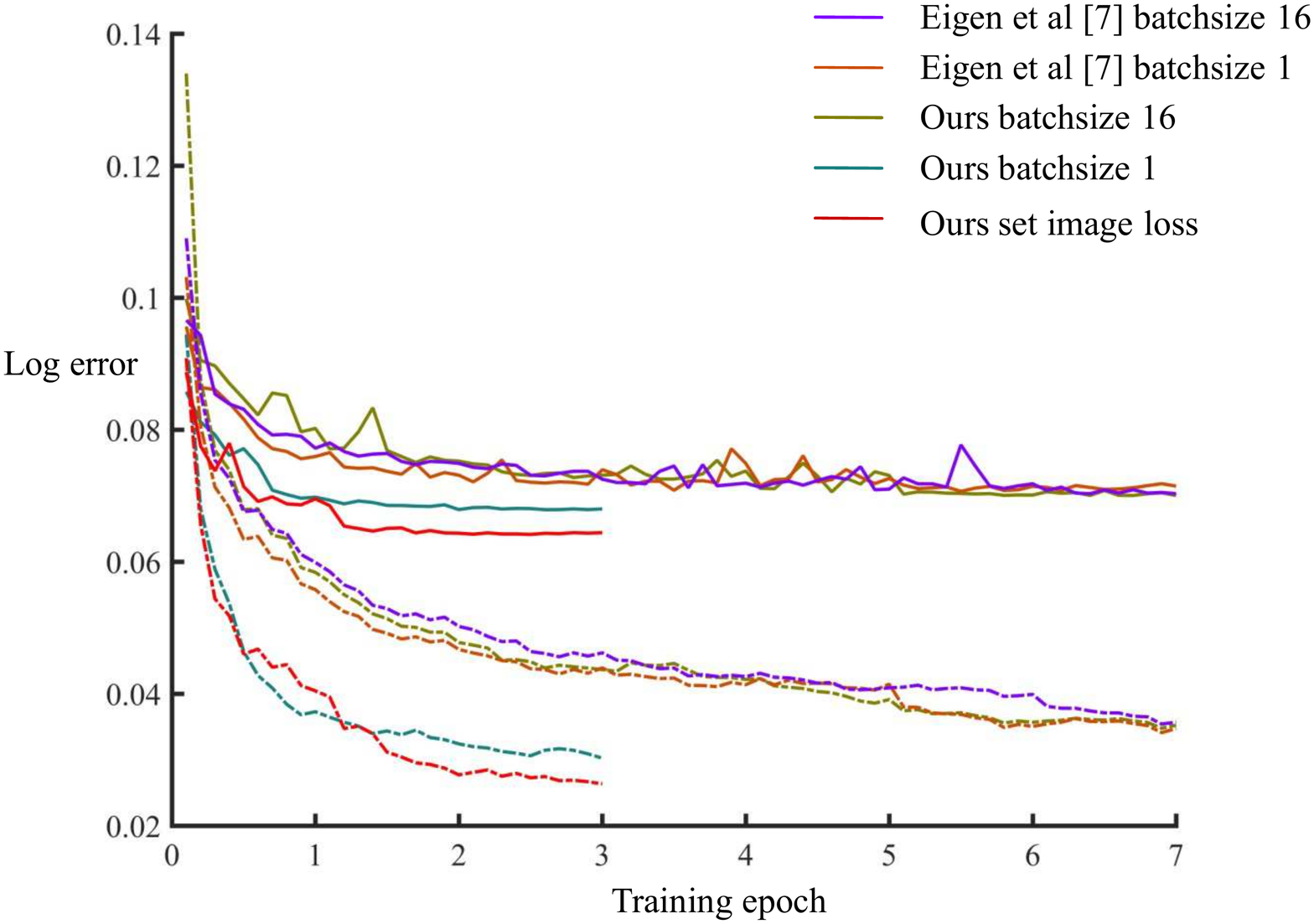}    
    \caption{A comparison of the $log_{10}$ training and test errors between our proposed network and~\cite{David15}. For clarity, we plot the $log$ error at every 0.1 epochs, and show only the first 7 epochs, even though the method of~\cite{David15} has not yet converged. The dashed lines denote training error, and solid lines denote testing error.  For our batch 1 and 16 results, we compare the errors for $L_{\text{single}}$ and $L_{\text{single}}$. 
    }
    \label{fig:convergence}    
    \vspace{-0.25cm}
\end{figure}

We show the convergence behaviour of our network for the joint training of the image parsing and feature fusion blocks
Figure~\ref{fig:convergence}.  Errors decrease faster with a batch size of 1 in comparison to a batch size of 16, and requires only 0.6M gradient steps or 2-3 epochs to converge. For the convergence experiment, we compare the single image loss (batch size 1, 16) with the set image loss as described in Section~\ref{sec:setloss} and observe that errors are lower with the set loss but convergence still occurs quickly.  Note that the fast convergence of our network is not due to the small batch size, but rather the improved architecture with the skip connection.  In comparison, the network architecture of~\cite{David15} requires a total of 2.5M  gradient steps to converge (more than 100 epochs) with batchsize 16, but even when trained with a batch size of 1, does not converge so quickly. Training for depth gradient estimates is even faster and converges within one epoch. Overall, our training time is about 70 hours (50 hours for depth learning and 20 hours for gradient learning)  on a single GPU TITAN X.


\section{Discussion and Conclusion}

We have proposed a fast-to-train multi-streamed CNN architecture for accurate depth estimation. 
To predict accurate and detailed depth maps, we introduced three novel contributions. First, we define a set loss jointly over multiple images. By regularizing the estimation between images in a common set, we achieve better accuracy than previous work. Second, we represent a scene with a joint depth and depth gradient representation, for which we learn with a two-streamed network, to preserve the fine detailing in a scene.  Finally, we propose two methods, one CNN-based and one optimization-based, for fusing the depth and gradient estimates into a final depth output.  Experiments on the NYU Depth v2 dataset shows that our depth predictions are not only competitive with state-of-the-art but also lead to 3D projections that are more accurate and richer with details. 

Looking at our experimental results as well as the results of state-of-the-art methods~\cite{Chakrabarti16,David15,laina2016deeper}, it becomes clear that the current numerical metrics used for evaluating estimated depth are not always consistent.  Such inconsistencies become even more prominent when the depth maps are projected into 3D. Unfortunately, the richness of a scene is often qualified by clean structural detailing which are difficult to capture numerically and which makes it in turn difficult to design appropriate loss or objective functions. Alternatively, one may look to applications using image-estimated depths as input, such as 3D model retrieval or scene-based relighting, though such an indirect evaluation may also introduce other confounding factors.  

Our method generates accurate and rich 3D projections, but the outputs are still only $111\!\times\!150$, whereas the original input is $427\!\times\!561$. Like many end-to-end applications, we work at lower-than-original resolutions to trade off the number of network parameters versus the amount of training data. While depth estimation requires no labels, the main bottleneck is the variation in scenes. The training images of NYU Depth v2 are derived from videos of only 249 scenes. The small training dataset size may explain why our set loss with its regularization term has such a strong impact. As larger datasets are introduced~\cite{dai2017scannet}, it may be become feasible to work at higher resolutions.  Finally, in the current work, we have addressed only the estimation of depth and depth gradients from an RGB source. It is likely that by combining the task with other estimates such as surface normals and semantic labels, one can further improve the depth estimates.

\paragraph{Acknowledgements:}  The authors would like to thank NVidia for their GPU donation to enable this research.





{\small
\bibliographystyle{ieee}
\bibliography{iccv17}
}

\end{document}